\documentclass{article}

\usepackage{anysize}
\marginsize{2.5cm}{2.5cm}{2.5cm}{2.5cm}

\usepackage[square,sort,comma,numbers]{natbib}
\usepackage{color}
\usepackage{caption}
\usepackage{subfigure}
\usepackage{comment}
\usepackage{amsmath}
\usepackage{amssymb}
\usepackage{wrapfig}

\usepackage{mathptmx}       
\usepackage{helvet}         
\usepackage{courier}        
\usepackage{makeidx}         
\usepackage{graphicx}        

\makeindex

\newcommand{\bs}[1]{\boldsymbol{#1}}

\renewcommand{\eqref}[1]{(\ref{#1})}

\newcommand{\s}[0]			{  {\bs{s}} }		
\newcommand{\z}[0]			{  {\bs{z}} }		
\newcommand{\y}[0]			{  {\bs{y}} }		
\newcommand{\x}[0]			{  {\bs{x}} } 
\renewcommand{\a}[0]			{ {\bs{\omega}} }

\newcommand{\pSA}[0]			{p(\s,\a)}
\newcommand{\qSA}[0]			{{q(\s,  \a )} }

\let \phiOld \phi
\let \PhiOld \Phi
\let \thetaOld \theta

\renewcommand{\phi}[0]			{\bs{\phiOld} }
\renewcommand{\Phi}[0]			{\bs{\PhiOld}}
\renewcommand{\theta}[0]			{\bs{\thetaOld}}

\definecolor{darkgreen}{rgb}{0,0.5,0}
\definecolor{orange}{rgb}{1,0.5,0}
\newcounter{cmt}

\begin{document}
\title{Learning Dynamic Robot-to-Human Object Handover from Human Feedback}
\author{Andras Kupcsik, David Hsu, Wee Sun Lee, \\
School of Computing, National University of Singapore, \\
\texttt{$\{$kupcsik,dyhsu,leews$\}$@comp.nus.edu.sg}}

\date{}
\maketitle

\abstract{ Object handover is a basic, but essential capability for robots
  interacting with humans in many applications, e.g., caring for the
  elderly and assisting workers in manufacturing workshops. It appears deceptively
  simple, as humans perform object handover almost flawlessly. The success of
  humans, however, belies the complexity of object handover as collaborative
  physical interaction between two agents with limited communication. This
  paper presents a
  learning algorithm for \emph{dynamic} object handover, for example, when a
  robot hands over water bottles to marathon runners passing by the water
  station. We formulate the problem as contextual policy search, in  which
  the robot learns object handover by interacting with the
  human. A key challenge here is to learn the latent reward of the handover task
  under \emph{noisy} human feedback. Preliminary experiments show that the
  robot learns to hand over a water bottle naturally and that it adapts
  to the dynamics of human motion. One challenge for the future is
  to combine the model-free learning algorithm with a model-based planning approach 
  and enable the robot to adapt over human preferences and object
  characteristics, such as shape, weight, and surface texture.}

\section{Introduction}

\begin{figure}[t]
\centering
\captionsetup{width=.8\linewidth}
\includegraphics[height = 1.5in]{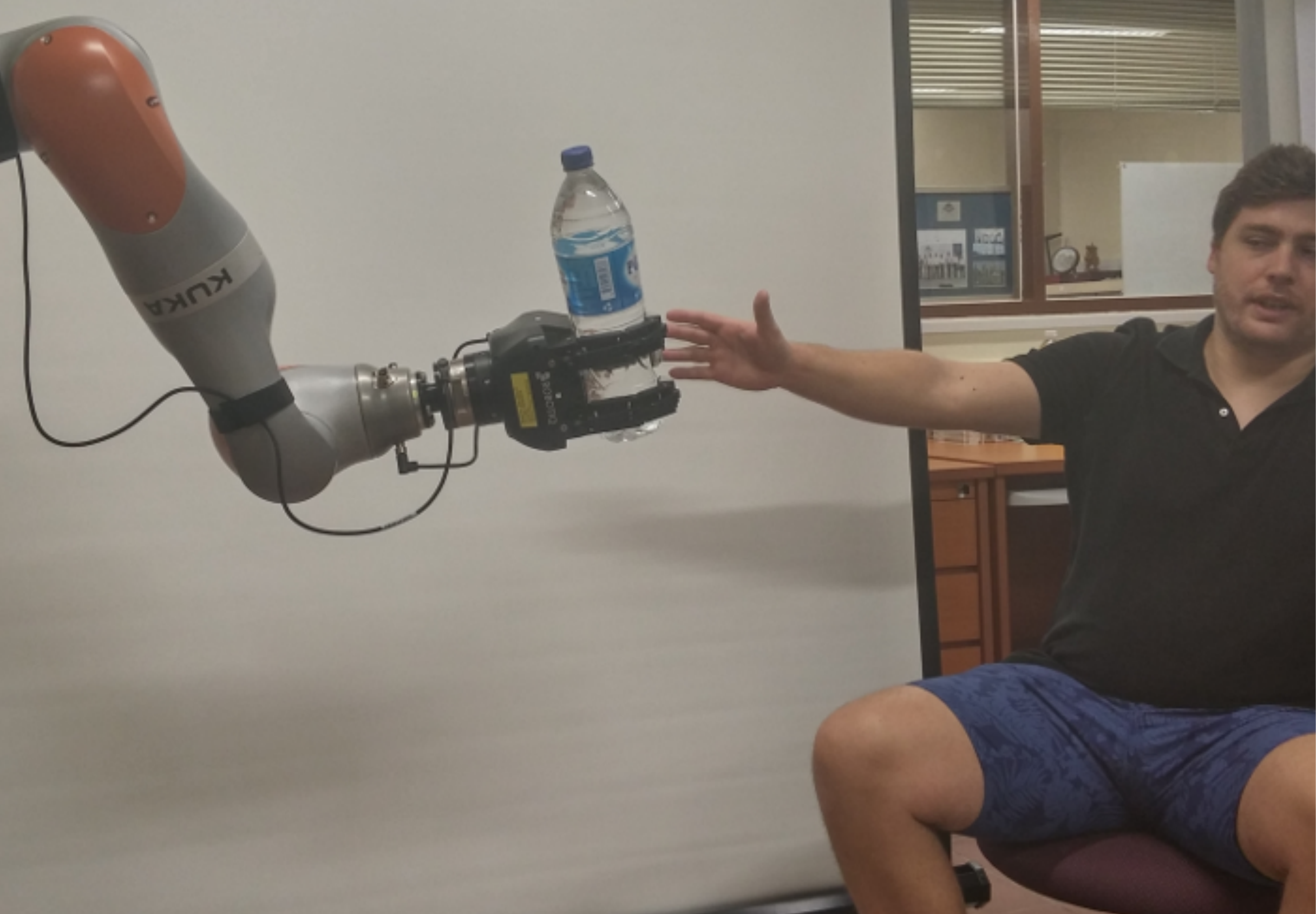}\quad
\includegraphics[height =1.5in]{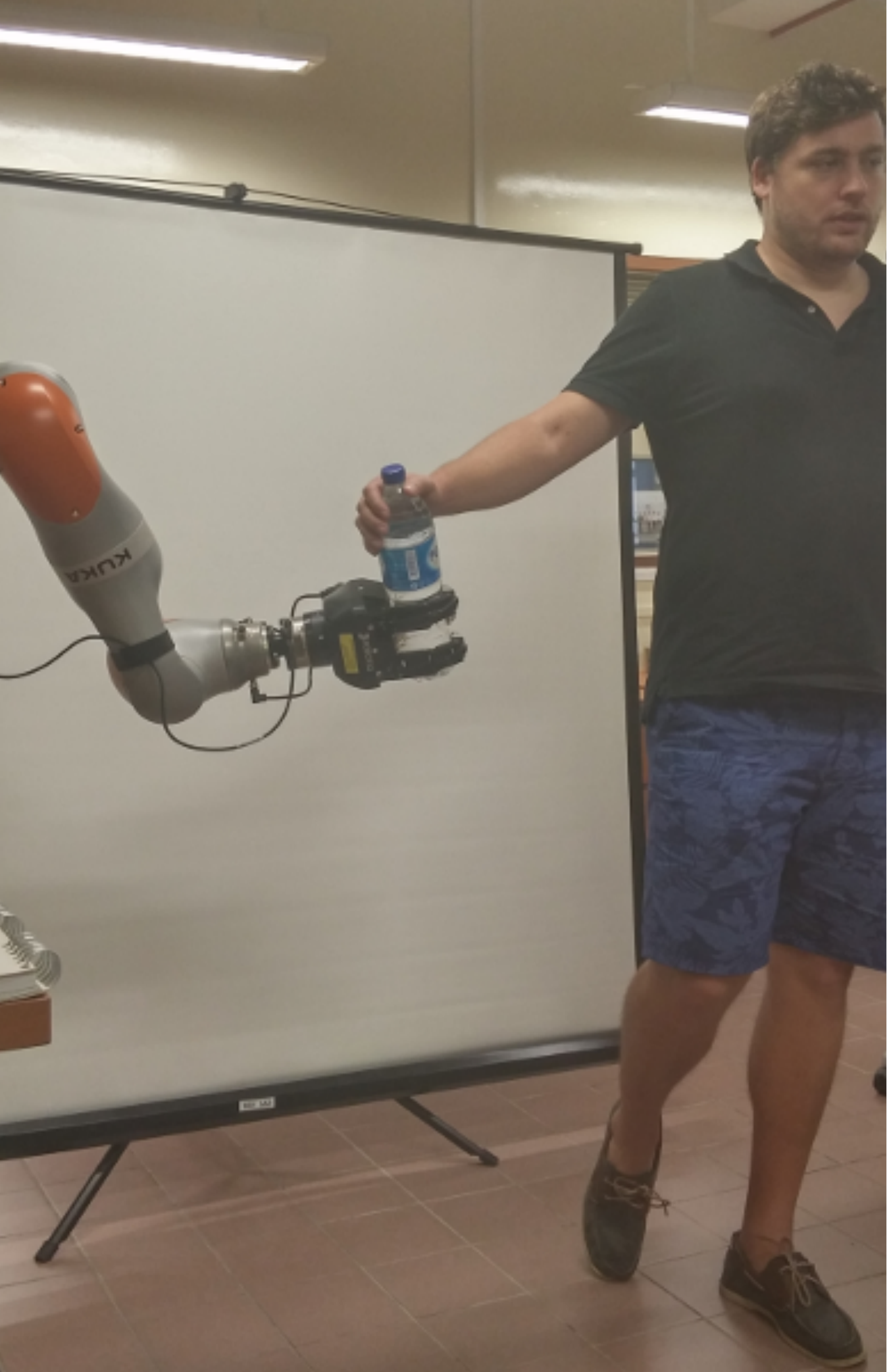}\quad
\includegraphics[height=1.5in]{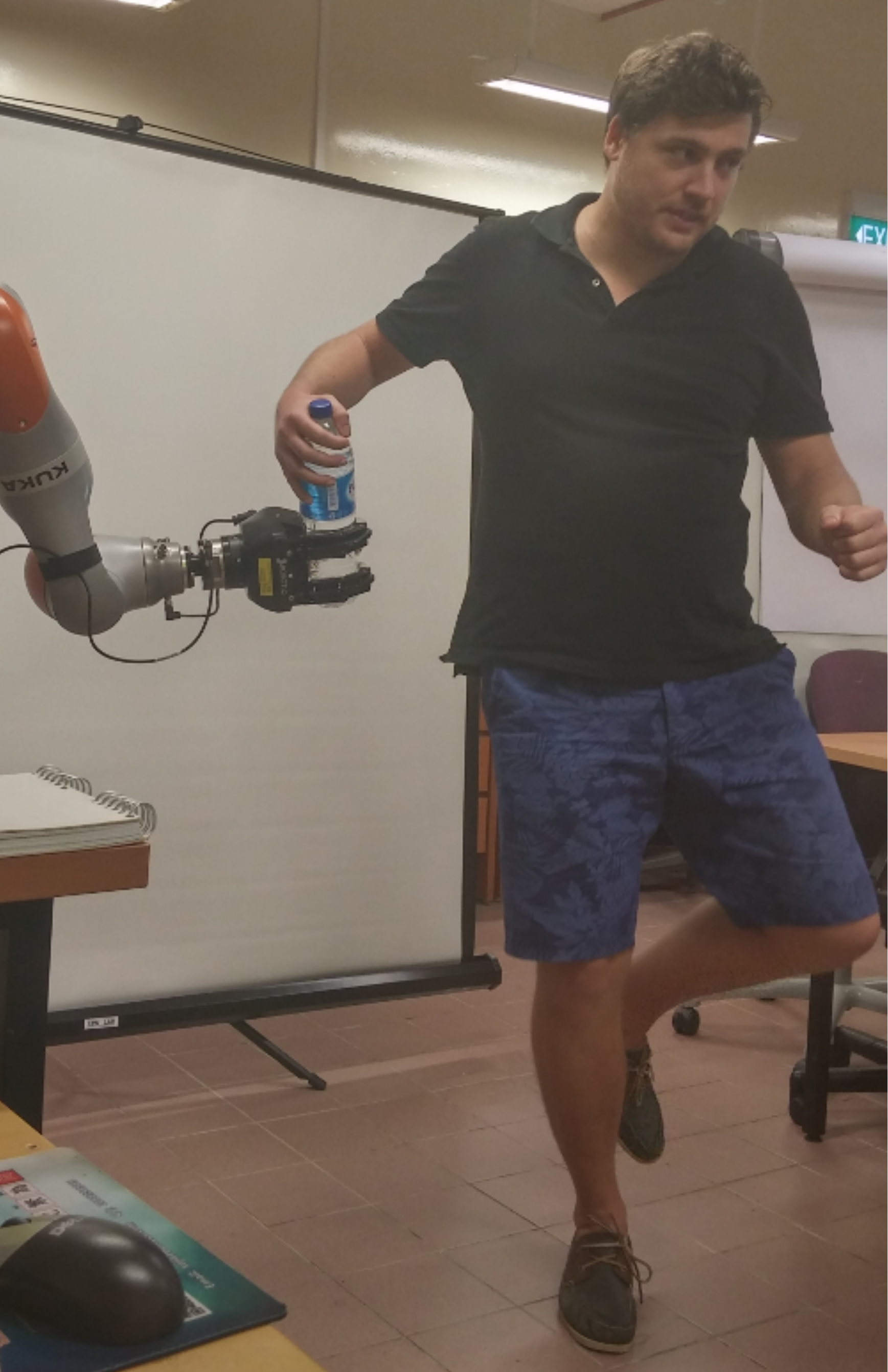}
\caption{Hand over a water bottle to a person sitting, walking, or running. }
\label{fig:handoverScenarios}
\end{figure}

In the near future, robots will become trustworthy helpers of humans,
performing a variety of services at homes and in workplaces.  A basic, but
essential capability for such robots is to fetch common objects of daily life,
e.g., cups or TV remote controllers, and hand them to humans. Today robots
perform object handover in a limited manner: typically the robot holds an
object statically in place and waits for the human to take it.  This is far
from the fluid handover between humans and is generally inadequate for the
elderly, the very young, or the physically weak who require robot services.  The
long-term goal of our research is to develop the algorithmic framework and the
experimental system that  enable robots to perform \emph{fluid} object handover in
a \emph{dynamic} setting and to adapt over human preferences and object
characteristics.
This work takes the first step and focuses on a robot handing over a water
bottle in a dynamic setting  (Fig.~\ref{fig:handoverScenarios}), e.g.,
 handing over flyers to people walking by or handing over  water bottles
to marathon runners.


Object handover appears deceptively simple. Humans are experts at object
handover. We perform it many times a day almost flawlessly without thinking 
and \emph{adapt} over widely different contexts:
\begin{itemize}
\item \emph{Dynamics:} We hand over objects to others whether they sit, stand, or
  walk by.
\item \emph{Object characteristics:} We hand over objects of different shape,
  weight, and surface texture.
\item \emph{Human preferences:} While typical human  object handover
  occurs very fast, we adapt our strategy and slow down when handing over
  objects to the elderly or young children.
\end{itemize}
The success of humans, however, belies the complexity of object handover as
collaborative  physical interaction between two agents with limited
communication.  Manually programming robot handover with comparable
robustness and adaptivity poses great challenge, as we lack even a 
moderately comprehensive and reliable  model for handover in a variety of contexts.

Alternatively, the robot can learn the handover skill by interacting with the
human and generalize from experience.  In this work, we formulate the learning
task as \emph{contextual policy search}~\citep{Kupcsik2013}. Policy search is a general
approach to reinforcement learning
and has been very successful in skill
learning for robot with many degrees of freedom~\citep{Deisenroth2013}.
Policy search algorithms
parametrize robot control policies and search for
the best parameter values by maximizing a reward function that captures the
policy performance.  Contextual policy search introduces a set of \emph{context
variables} that depend on the task context, e.g., object type or size for the
handover task, and the policy parameters are conditioned on the context
variables.

A reward function that accurately measures policy performance is key to the
success of policy search. However, handcrafting a good reward function is often
tedious and error-prone, in particular, for learning object handover. It is
unclear what quantitative measures capture fluid object handover. Instead, we
propose to learn the latent reward function from human feedback.  Humans are
experts at object handover and can easily provide reward feedback. However, the
feedback is often noisy. To be robust against noise and avoid overfitting, we
apply a Bayesian optimization approach to latent reward learning. Importantly,
our learning algorithm allows for both \emph{absolute feedback}, e.g., ``Is
the handover good or bad?'', and \emph{preference feedback} , e.g., ``Is the
handover better than the previous one?''. Combining latent reward learning and
policy search leads to a holistic contextual policy search algorithm that
learns object handover directly from human feedback. Our
preliminary experiments show that the
  robot learns to hand over a water bottle naturally and that it adapts
  to the dynamics of human motion.

\section{Related Work}
\label{sec:related}

\subsection{Object Handover}
Object handover has intrigued the research community for a long time from the
both physical and social-cognitive perspectives. Early work on handover dates
back to at least 1990s~\citep{Agah_ICRA_1997,NagOos98}.  Recent work
suggests that object handover consists of three stages conceptually: approach,
signal, and transfer~\citep{Strabala_JHRI_2013}.  They do not necessarily
occur sequentially and may partially overlap.  In the first stage, the giver
approaches the taker and poses the object to get ready for
handover~\citep{CakSri11,MaiGha12,SisAla05}. In the second stage, the giver
and taker signal to each other and exchange information, often through
non-verbal communication, such as motion~\cite{Dragan_RSS_2013}, eye gaze, or
head orientation~\citep{Grigore_IROS_2013}, in order to establish joint
intention of handover.  In the final stage, they complete the physical
transfer of the object. The transfer stage can be further divided into two sub-stages,
before and after the giver and the taker establish joint contact of the
object, respectively. Earlier work on object transfer generally assumes that
the object remains stationary once joint contact is established and relies on
handcrafted
controllers~\citep{Agah_ICRA_1997,Chan_ICRA_2014,HuaCak15,NagOos98}.  Our
work focuses to the final physical transfer stage only. The algorithm learns a
controller directly from human feedback. It does not make the stationary
assumption and caters for dynamic handover.  Object transfer is an instance of
the more general problem of cooperative manipulation~\citep{BruKha08}: it
involves two asymmetric agents with limited communication.

Human-human object handover provides the yardstick for handover performance.
Understanding how humans perform handover (e.g., \citep{ChaPar13,HubKup13})
paves the way towards improved robot handover performance.

\subsection{Policy Search}
Robot skill learning by policy search has been highly successful in recent
years \cite{Deisenroth2013}. Policy search algorithms learn a skill represented
as a
probability distribution over parameterized robot controllers, by 
maximizing the expected reward.
To allow robot skills to adapt to different situations, contextual policy
search learns a contextual policy that conditions a skill on context
variables~\citep{Silva2012, Daniel2012, Kupcsik2013}.

To represent robot skills, policy search 
typically makes use of  parametrized controllers, such
as \emph{dynamic movement primitives}~\citep{Ijspeert2003}
or \emph{interaction primitives} \citep{Heni_ICRA_2014}. The latter 
is well-suited for human-robot interaction tasks.
Our work, on the other hand, exploits domain knowledge to construct a
parameterized impedance controller. 

To learn robot skills, policy search requires that a reward function
be given to measure learning performance.  However, 
handcrafting a good reward function is often difficult.
 One approach is
inverse reinforcement learning (IRL), also called inverse optimal control,
which learns a reward function from expert demonstration~\cite{Ng_ICML_2000,
  Ratliff_2009_AR}.  Demonstrations by human experts can be difficult or tedious to
acquire, in particular, for robot-human object handover.
An alternative is to learn  directly from human feedback,
without human expert demonstration.  Daniel et al. use reward feedback from
humans to learn manipulation skills for robot
hands~\citep{Daniel_RSS_2014}. Wilson et al.  consider learning control
policies from trajectory preferences using a Bayesian approach without
explicit reward feedback~\citep{Wilson_NIPS_2012}.  Jain et al. learn
manipulation trajectories from human preferences \citep{Jain_CoRR_2013}.
Preference-based reinforcement learning algorithms generally do not use
absolute reward feedback and rely solely on preference feedback
\citep{Wirth_ECML_2013}.
 Our algorithm 
combines both absolute  and preference feedback in a single
Bayesian framework to learn a reward
function and integrate with policy search for robot skill learning.

\section{Learning Dynamic Handover from Human Feedback}
\label{sec:handover}
\subsection{Overview}
Assume that a robot and a human have established the joint intention of
handover.  Our work addresses the physical transfer of an object from the
robot to the human.
The robot controller $u(\cdot\, ; \a )$ specifies the control action $u_t$ at
the state $\x_t$ at time $t$ for $t=1, 2, \dots$. The controller $u(\cdot\, ; \a )$
is parametrized by a set of parameters $\a$, and the notation makes the
dependency on $\a$ explicit. A reward function  $R(\a)$ assigns a real number
that measures the performance of the policy $u(\cdot\, ; \a )$.
To handle the dynamics of handover, we introduce a context
variable $\s$ representing the velocity of the human hand and condition the
controller parameters $\a$ on $\s$, giving rise the reward function $R(\a, \s)$. 
In general, context variables may include other features, such as human
preferences and object characteristics as well.
A contextual policy $\pi(\a | \s)$
is a probability distribution over parametrized controllers, conditioned on the
context $\s$. Our goal is to learn a contextual policy that maximizes the
expected reward:
\begin{align}
\pi^* = \arg\max_{\pi} \int_{\s} \int_{\a} R(\a, \s) \pi(\a|\s) \mu(\s)\;
\mathrm{d}\a \,\mathrm{d}\s, 
\end{align}
where $\mu(\s)$ is a given prior distribution over the contexts.

\begin{figure}[t]
\centering
\captionsetup{width=.8\linewidth}
\includegraphics[width = 8cm]{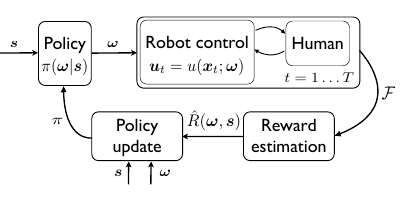}
\caption{The human-robot handover skill learning framework. The robot observes
context $\s$, then samples $\a$ using the policy $\pi(\a|\s)$. The experiment is
executed with a robot controller with parametrization $\a$. The robot controller
$u(\x;\a)$ provides deterministic control signals $\bs{u}$ given the state
of the robot and its environment $\bs{x}$. After the experiment the human
provides a high-level feedback $\mathcal{F}$, which is used the estimate the
latent reward $\hat{R}(\a,\s)$. Finally, the policy is updated with the latest
data.}
\label{fig:hri-setup}
\end{figure}

Contextual policy search iteratively updates $\pi$ so that the distribution
peaks up on controllers with higher rewards.  In each iteration, the robot
learner observes context~$\s$ and samples a controller with parameter value
$\a$ from the distribution $ \pi(\cdot |\s)$.  It executes the controller
$u(\cdot | \a)$ and observes the reward $R(\a,\s)$.  After repeating this
experiment $L$ times, it updates $\pi$ with the gathered data
$\{\a_i,\s_i,R(\a_i,\s_i)\}_{i=1}^L$ and proceeds to the next iteration.
See Fig.~\ref{fig:hri-setup} for the overall learning and control
architecture and Table~\ref{alg:handover}
for a sketch of our learning algorithm.

The reward function $R(\a, \s)$ is critical in our algorithm.
Unfortunately, it is difficult to specify manually a good reward function for learning
object handover, despite the many empirical studies of human-human object
handover~\citep{CakSri11, ChaPar13,HubKup13, Strabala_JHRI_2013}.  We
propose to learn a reward function $\hat{R}(\a, \s)$ from human
feedback. Specifically, we allow both \emph{absolute} and \emph{preference}
human feedback. Absolute feedback provides direct assessment of the robot
controller performance on an absolute scale from 1 to 10.
Preference feedback compares one controller with another relatively.
While the former has higher information
content, the latter is usually easier for humans to assess.
We take a Bayesian approach and apply Gaussian process regression to latent
reward estimation.  The learned reward model $\hat{R}(\a, \s)$ generalizes the
 human feedback data. It provides estimated reward on arbitrarily sampled
$(\a,\s)$ without additional experiments and drastically reduces the
number of robot experiments required for learning a good  policy.

\begin{table}
\centering
 \begin{tabular}{ p{0.8\textwidth} }
\hline
The C-REPS Algorithm with Human Feedback \\
\hline
\textbf{Input:} relative entropy bound $\epsilon$, initial policy $\pi (\a|\s)$, 
maximum number of policy updates $H$. \\
\textbf{for} $h = 1,\dots,H$ \\
\quad\begin{tabular}{| p{90mm} } 
      \textbf{Collect human feedback data:} \\
     \quad\begin{tabular}{| p{\textwidth} }
	{\it Observe context $\s_i\sim \mu(\s)$, $i = 1,\dots,L$}\\
    {\it Draw parameters $\a_i \sim \pi(\a| \s_i)$}\\
    {\it Collect human feedback $\mathcal{F}_i$}\\
     \end{tabular}
     \textbf{Estimate latent rewards of all previously seen samples} $\{\a_i, \s_i, \mathcal{F}_i\}_{i=1}^{E}$  \\
       \textbf{Predict rewards:} \\
       \quad\begin{tabular}{| p{.9\textwidth} } 
         {\it Draw context  $\s_j \sim \hat{\mu}(\s),~j=1,\dots,Q$}\\
         {\it Draw parameters $\a_j \sim \pi(\a|\s_j)$}\\
         {\it Predict $\hat{R}(\a_j,\s_j)$ with reward model } \\
		\end{tabular}
   \textbf{Update policy:} \\
   \quad\begin{tabular}{| p{.9\textwidth} }
     {\it Update policy $\pi(\a|\s)$ using C-REPS with samples $\{\a_j, \s_j,
	 \hat{R}(\a_j, \s_j)\}_{j=1}^Q$}
     \end{tabular}
     \textbf{end} \\
 \end{tabular} \\
\textbf{Output:} policy $\pi(\a| \s)$ \\
\hline
\caption{The learning framework for human-robot object transfer.}
\label{alg:handover}
\end{tabular}
\end{table}

\subsection{Representing the Object Handover Skill}

	\label{sec:skill}
In this section we discuss how we encode the handover skill and which parameters $\a$
refers to. In our work we use a trajectory generator, a robot arm controller and a robot hand controller to encode the handover skill. 
A trajectory generator provides reference Cartesian coordinates for the robot
end-effector to follow. In robot learning tasks, Movement Primitives (MP) are
often used to encode a trajectory with a limited amount of parameters. 
MPs encode the shape, speed and magnitude of the trajectory in Cartesian space, or in joint space for each degree of freedom. 
While MPs can encode a wide variety of skills, they typically require a higher number
of parameters to tune, which might slow down the learning process.

For handover tasks however, we can use human expert knowledge to define robot hand
trajectories. This approach allows for a more compact representation of the
trajectory generator with less parameters to tune. Furthermore, we can address
safety by reducing the workspace of the robot and we can easily synchronize with
the human motion. In our experiments we use visual data
of a Kinect sensor, which tracks the right hand of the human. As soon as the human
hand is within $d_{max}$ distance from the robot hand the robot moves the object towards the human hand
location. We assume that a path planner computes the reference trajectory from
the current robot hand location to the human hand location. The reference
trajectory  
is updated every time the human hand location is updated. As soon
as the distance between the human and the robot hand falls below $d_{min}$, we
do not use visual information due to possible occlusion and measurement error.
	Instead, we use the recorded visual data to predict the human hand
	trajectory for the next second when the physical interaction is likely to
	happen.  The values of $d_{min}$ and $d_{max}$ may depend on different
	factors, such as, experiment setup, robot configuration, etc.

In order to ensure robust
human-robot handover, we need to allow compliant robot arm motion. 
We use Cartesian impedance control \citep{BruKha08}
where the wrench $\bs{F}_{6 \times 1}$
concatenating forces and torques exerted in the end-effector frame is
computed according to
$
\bs{F} = \bs{M} \Delta \bs{\ddot{\x}} + \bs{D} \Delta\bs{\dot{\x}} + \bs{P}
\Delta\x,
$
where $\Delta\x_{6\times 1}$ is the deviation from the reference trajectory.
The gain parameters $\bs{M}$, $\bs{D}$ and
$\bs{P}$ will determine the amount of exerted forces and torques. 
$\bs{M}$ is typically replaced with the robot inertia at the current
state. We choose the damping $\bs{D}$ such that the 
closed loop control system is critically damped. We use a diagonal stiffness matrix
$\bs{P} = \mbox{diag}([\bs{p}_t^T, \bs{p}_r^T])$, where $\bs{p}_t$
is the translational and $\bs{p}_r$ is the rotational stiffness. Finally, the
applied torque commands are $\bs{\tau} = \bs{J}^T \bs{F} + \bs{\tau}_{ff}$,
where $\bs{J}$ is the Jacobian of the robot and $\bs{\tau}_{ff}$ are feed forward
torques compensating for gravity and other nonlinear effects.

Motivated by recent work in human-human handover experiments \cite{ChaPar13}, a robot grip force
controller \citep{Chan_ICRA_2014} has been proposed
$
\bs{F}_g = k \bs{F}_l + \bs{F}_{ovl}, 
$
where $\bs{F}_g$ is the commanded grip force, $\bs{F}_l$ is the measured load
force and $\bs{F}_{ovl}$ is the preset overloading force. The slope 
parameter $k$ depends on object properties, such as mass, shape
and material properties. When using this controller, the robot will release the
object in case the total load force on the robot drops below a threshold value.
For robot hands with only finger position control we cannot use the above
control approach. Instead, we directly command finger positions by identifying
the finger position with minimal grip force that still holds the object. Then,
we use a control law to change finger positions linear in the load force
$\bs{f}_{pos}= \bs{f}_{min} + m \bs{F}_l$. The value of $m$ depends on many
factors, such as, object type, weight and other material properties.

For learning the object handover, we tune $7$ parameters of the
control architecture. For trajectory generator we tune the minimal and maximal
tracking distances $d_{min}$ and $d_{max}$. For the compliant arm controller we learn the translational stiffness
parameters and one parameter for all the rotational stiffness values. Finally,
for finger controller we tune the slope parameter. All these parameters are
	collected in $\a_{7\times 1}$.

\subsection{Estimating the Latent Reward Function}

In this section we propose a Bayesian latent reward estimation technique based on previous work \citep{Chu_ICML_2005}.
Assume that we have observed a set of samples $\{\s_i, \a_i\}_{i=1}^E$ and human
feedback $\{\mathcal{F}_i\}_{i=1}^E$, where $\mathcal{F}_i = \tilde{R}(\y)$, in
case the human gives an absolute evaluation (denoted by $\tilde{R}$) on
parametrization $\a_i$ in context $\s_i$, $ \y= [\s^T, \a^T]^T$. In case of
preference feedback $\mathcal{F}_i = \y_k \succ \y_{i\neq k}$ if $\y_i$ is
preferred over $\y_i$. Note that for a given sample there may exist both
preference and absolute evaluation.

We define the prior distribution over the latent rewards as a Gaussian Process
\citep{Rasmussen2006}, $\hat{\bs{R}} \sim \mathcal{N}(\bs 0, \bs K)$, with
$\bs{K}_{ij} = k(\y_i, \y_j)$. Without the loss of generality we assume $\bs{0}$
prior mean, but more informative priors can be constructed with expert
knowledge. The likelihood function for preference based feedback is given by
$p(\y_i \succ \y_j|\hat{\bs{R}}) = \bs \Phi((\hat{R}_i -
\hat{R}_j)/(\sqrt{2}\sigma_p))$ \citep{Chu_ICML_2005}, where $\bs{\Phi}(\cdot)$
is the c.d.f. of $\mathcal{N}(0,1)$ and $\sigma_p$ is a noise term accounting
for feedback noise. For absolute feedback data we simply define
the likelihood by $p(\tilde{R}_i|\hat{\bs{R}}) = \mathcal{N}(\hat{R}_i,
\sigma_r^2)$, where $\sigma_r^2$ represents the variance of absolute human
feedback. Finally, the posterior distribution of the latent rewards can be
approximated by,
\begin{equation}
p(\hat{\bs{R}}|\mathcal{D}) \propto \prod_{i=1}^N p(\y_{i,1} \succ \y_{i, 2}|\hat{\bs{R}}) \prod_{j=1}^M p(\tilde{{R}}_j|\hat{R}_j,\sigma_r^2) p(\hat{\bs{R}}|\bs 0, \bs K),
\end{equation}
where we used the notation $p(\y_{i,1}\succ \y_{i,2}|\hat{\bs{R}})$ to highlight that $\mathcal{F}_i$ is a preference feedback comparing $\y_{i,1}$ to $\y_{i,2}$. For finding the optimal latent rewards, we minimize 
\begin{equation}
J(\hat{\bs{R}}) = - \sum_{i=1}^N \log \bs \Phi (\z_i) +  \frac{\sigma_r^{-2}}{2} \sum_{j=1}^M (\tilde{R}_j - \hat{R}_j)^2 + \hat{\bs{R}}^T \bs K^{-1} \hat{\bs{R}}, \label{eq:optpref}
\end{equation}
with $\z_i = (\hat{R}(\y_{i,1}) - \hat{R}(\y_{i,2}))/(\sqrt{2}\sigma_p)$. It was
shown in \citep{Chu_ICML_2005} that minimizing $J$ w.r.t. $\hat{\bs{R}}$ is a
convex problem in case there is only preference based feedback ($M=0$). However,
it easy to see that the Hessian of $J(\hat{\bs{R}})$ will only be augmented with non-negative elements in the diagonal in case $M> 0$, which will leave the Hessian positive semi-definite and the problem convex. Optimizing the hyper-parameters of the kernel function $\bs \theta$ and the noise terms can be evaluated by maximizing the evidence $p(\mathcal{D}|\bs \theta, \sigma_p, \sigma_r)$. While the evidence cannot be given in a closed form, we can estimate it by Laplace approximation.

It is interesting to note that in case there is only preference feedback, that
is, $M=0,~N>0$, we obtain the exact same algorithm as in \citep{Chu_ICML_2005}.
In the other extreme, in case there is only absolute feedback ($M>0,~N=0$) we
get Gaussian Process regression, which provides a closed form solution for
$p(\hat{\bs{R}})$. Overall, our extension provides an opportunity to mix preference and absolute feedback in a unified Bayesian framework. 

Also note that after obtaining $p(\hat{\bs{R}})$ we can use Bayesian linear
regression to query the expected reward $R^*$ of unseen samples $\bs{y}^*$
\cite{Chu_ICML_2005, Rasmussen2006}. We can use the resulting generative model
of the reward to query the reward for a large number of samples from the current
control distribution $\y \sim \mu(\s)\pi(\a|\s)$, without the need for real
experimental evaluation. Such a data-efficient model-based approach has been
demonstrated to reduce the required number of experiments up to two orders of
magnitude  \citep{Kupcsik2013,Daniel_RSS_2014}. 

\subsection{Contextual Relative Entropy Policy Search}

To update the policy $\pi(\a|\s)$, we rely on the contextual extension of Relative Entropy Policy Search \citep{Kupcsik2013,Deisenroth2013}, or C-REPS.
The intuition of C-REPS is to maximize the expected reward over the joint
context-control parameter distribution, while staying close to the observed data to balance  out exploration and experience loss.
C-REPS uses an information theoretic approach, where the relative entropy between consecutive  parameter distributions is bounded
$ \int_{\s,\a} p(\s, \a) \log 
 \frac{p(\s, \a)}{q(\s,\a)}d\s d\a \leq \epsilon ,
 $
   where $p(\s,\a)$ and $q(\s, \a)$ represent the updated and the previously used context-parameter distributions. The parameter         $\epsilon \in \mathbb{R}^+$ is the upper bound of the relative entropy. The emerging constrained optimization problem can be solved by the Lagrange multiplier method (see e.g. \citep{Kupcsik_AIJ_2015}). The closed form solution for the  new distribution is given by 
$
    \pSA \propto \qSA \exp\left((R(\a, \s) - V(\s))/\eta\right).
 $
  Here, $V(\s)$ is a context dependent baseline, while $\eta$ and $\bs{\theta}$
  are Lagrangian parameters. The baseline is linear in some context features and
  it is parametrized by $\bs{\theta}$. To update the policy we use the computed
  probabilities $\pSA$ as sample weights and perform a maximum likelihood
  estimation of the policy model parameters.

\section{Experiments}
\label{sec:expr}

For the handover experiment we use the 7-DoF KUKA LBR arm (Figure 
3). 
 For the robot hand we use the Robotiq
	3-finger hand. The fingers are position controlled, but the maximum grip force can
	be indirectly adjusted by limiting the finger currents. In order for
	accurate measurement of external forces and torques, a wrist mounted
	force/torque sensor is installed.

\begin{figure}[t]
\centering
\captionsetup{width=.8\linewidth}
\includegraphics[width = 0.6\textwidth]{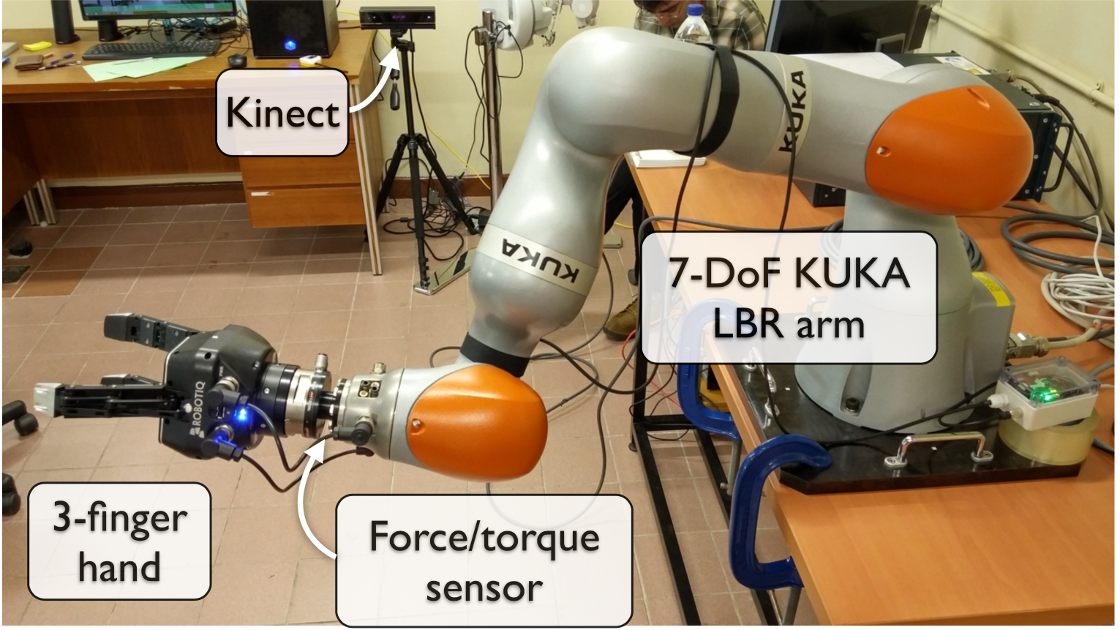}
\begin{center}
\caption{Robot setup for experiments. We use the 7-DoF KUKA LBR arb with the
3-finger Robotiq robot hand. We use Kinect to track the human hand motion. }
\end{center}
\label{fig:experiment}
\end{figure}

\subsection{Experimental Setup}

An experiment is executed as follows. First, a 1.5l water bottle is placed at a
fixed location, which the robot is programmed to pick up. Subsequently, the
robot moves the bottle to a predefined position. At this point we
enable compliant arm control and we use a Kinect sensor (Figure 3) to track the hand of the human. Subsequently, the human
moves towards the robot to take the bottle. While approaching the robot, we use
the Kinect data to estimate the hand velocity $\s$ of the human, which we assume
to be constant during the experiment. We only use data when the human is
relatively far (above $1$m) from the robot to avoid occlusion. After the context variable is
estimated the robot sets its parameter by drawing a controller parametrization
$\a \sim \pi(\a|\s)$. Subsequently, the robot and the human make physical
contact and the handover takes place. Finally, the human evaluates the robot
performance (preference or absolute evaluation on a 1-10 scale, where 1 is worst
10 is best) and walks away such that the next experiment may begin. 

We presented the pseudo code of our learning algorithm in Table \ref{alg:handover}. 
As input to the algorithm we have to provide the initial policy $\pi(\a|\s)$,
and several other parameters.
We use a Gaussian distribution
to represent the policy $\pi(\a|\s) = \mathcal{N}(\a|\bs{a} + \bs{A} \s,
\bs{\Sigma})$. In the beginning of the learning we set $\bs{A}=\bs{0}$, that is,
the robot uses the same controller distribution over all possible context values. During learning
all the parameters ($\bs{a},~ \bs{A},~ \bs{\Sigma}$) of the policy will be tuned according to the C-REPS update rule.

\begin{figure}
\centering
\captionsetup{width=.8\linewidth}
\includegraphics[width = 0.3 \textwidth]{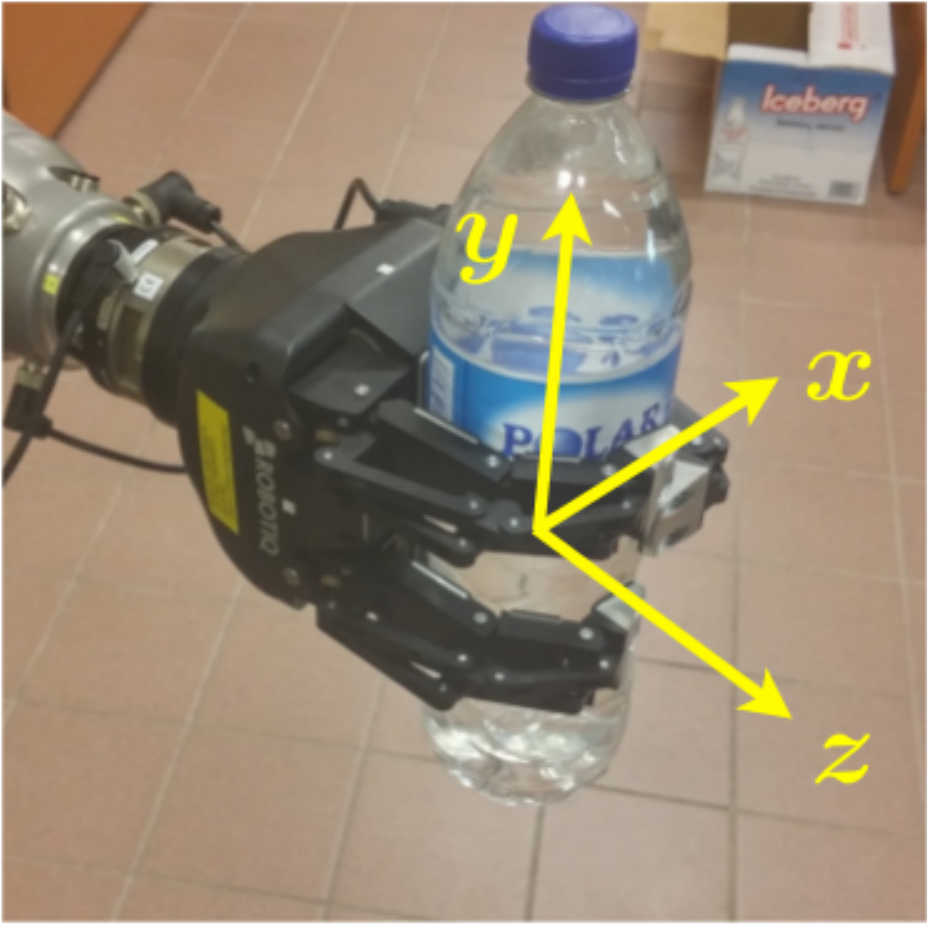}
\caption{The robot hand frame orientation. }
\label{fig:robothand}
\end{figure}

The initial policy mean $\bs{a}$ and the diagonal elements of the covariance
matrix $\bs{\Sigma}$ are set as follows. For the rotational stiffness we set $2.75$ Nm/rad mean and $0.5^2$
variance. For the translational stiffness parameters we chose $275,~450,~275$
N/m in x, y, and z direction in the hand frame (Fig \ref{fig:robothand}). The
variances are $50^2, 75^2$, and $50^2$ respectively. For the finger control slope parameter we
chose $2.5$ 1/N with a variance of $0.5^2$. This provides a firm grip of the
water bottle. The robot will not move the fingers until the force generated by
the human hand reaches half the weight of the bottle. With a slope parameter of
$0$ the robot exerts a minimal grip force that can still support the bottle.
With a slope value above $5$ the robot only releases the bottle if the human can
support $1.2\times$ the object weight. Thus, we can avoid
dropping the object, even with the initial policy. 
Finally as mean we set $200$mm and
$600$mm as minimal and maximal trajectory tracking control distance. As
variances we chose $50^2$ and $150^2$. 
The parameters are therefore initialized as $\bs{a} = (2.75,~ 275,~
450,~275,~2.5,~200,~600)^T$, $\bs{A} = \bs{0}$ and $\bs{\Sigma} =
\mbox{diag}(0.5^2,~ 50^2,~75^2,~50^2,~0.5^2,~50^2,~150^2)$.

For the C-REPS learning algorithm in Table \ref{alg:handover} we chose
$\epsilon = 0.75$ and we updated the policy after evaluating $L = 10$
human-robot handover experiments.
However, before the first policy update we used $L = 40$ handover experiments, such that
we have a reliable estimation of the latent rewards. Before each policy update
we estimate the latent rewards for all the previously seen experiments $\{\a_i, \s_i,
\mathcal{F}_i\}_{i=1}^E$. Here, $E$ represents the total number of observed
samples. Note, that $E$ is increased by the amount of latest experiments $L$
before each policy update. Therefore, $E$ represents how much experimental
evaluation, or information we used to reach a certain level of performance.
After estimating the latent rewards we use the resulting generative reward model
to evaluate $Q=500$ \emph{artificial} context-control parameter pairs drawn from  
$\hat{\mu}(\s)\pi(\a|\s)$.  We
used these artificial samples to update the policy. This way we got a highly
data-efficient algorithm, similar to the one in \citep{Kupcsik2013}. After the
policy is updated, we start a new loop and evaluate $L$ new experiment. 
We not only use this information to update our dictionary to
estimate latent rewards, but also to estimate the performance of the current
policy. The performance of the policy is measured by the expected latent reward of the
newly evaluated $L$ experiments. We expect the performance measure to increase 
with the amount of information $E$ and policy updates.
After updating the policy $H$ times (Table \ref{alg:handover}) we terminate the learning.

\subsection{Results}

\begin{figure}
\centering
\captionsetup{width=.8\linewidth}
\includegraphics[width = 0.4 \textwidth]{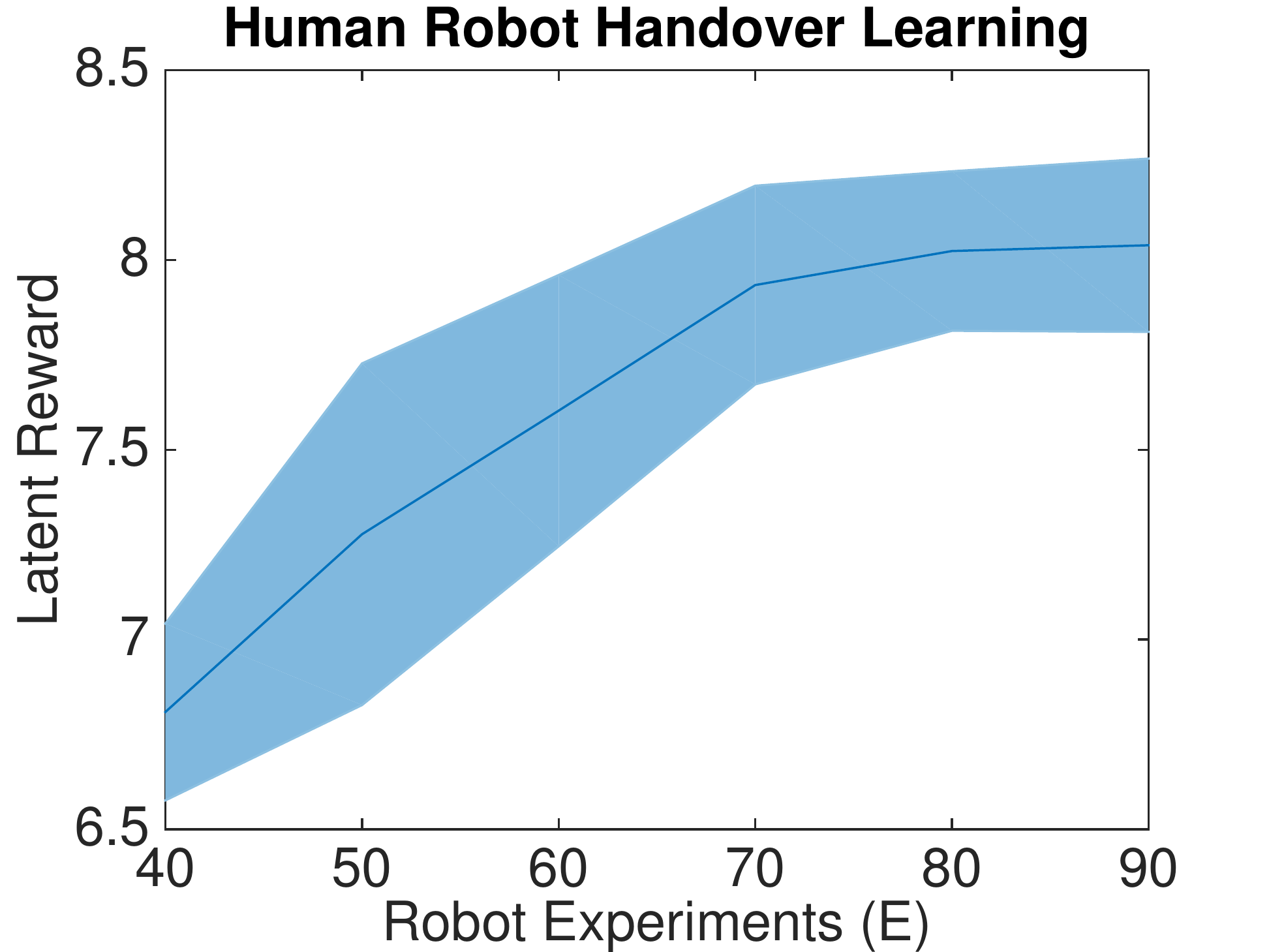}
\caption{The expected latent reward mean and standard deviation over $5$
independent learning trials. Humans may give absolute feedback in a 1 to 10 scale. Initially the latent reward is estimated to be around $6.8$, which goes up to around $8$ after evaluating the learning.}
\label{fig:learningResults}
\end{figure}

As the learning algorithm uses randomly sampled data for policy updates and
noisy human feedback, the
learned policy and its performance may vary.
 In order to measure the consistency of
the learning progress we repeated the complete learning trial several times. A
trial means evaluating the learning algorithm starting with the initial policy and with an empty
dictionary, $E=0$, but using the same parameters for $L$ and $\epsilon$. 
We evaluated $5$ learning trials and recorded the expected performance of the
robot before each policy update. The expected learning performance over $5$
trials with 95$\%$ confidence bounds against the amount of real
robot experiments $E$ used for policy update is shown in Figure
\ref{fig:learningResults}. We can see that learning indeed improved the
performance of the initial policy, which has an expected value of $6.8$. Over
the learning trials, we noticed
that the human mostly gave absolute feedback for very good or bad solutions. This
is expected as humans can confidently say if a handover skill feels close to
that of a human, or if it does something unnatural (e.g., not releasing the
object). By the end of the learning, 
the expected latent reward rose to the region of $8$. Note, that the variance of
the learning performance over different trials not only depends on the
stochastic learning approach, but also on noisy human feedback. 
Thus we can conclude that the learning indeed
improved the expected latent reward of the policy, but how did the policy and
the experiments change with the learning?

\textbf{The learned policy.} We first discuss the mean value $a$ of the learned
policy and then we show how the policy generalizes to more dynamic tasks.
Over several learning trials we 
observed that a high quality policy provides a lower rotational stiffness
compared to the hand-tuned initial policy. We observed that on expectation the
learned rotational stiffness is $1.29$ Nm/rad, which is lower than the initial
$2.75$. This helped the robot to quickly orient the object with the human hand
upon physical contact. We observed similar behavior in the translational
stiffness values in the $x-z$ directions (see Figure \ref{fig:robothand}). The
learned values were almost $100$ N/m lower compared to the initial values. This
helps the robot to become more compliant in horizontal motions. Interestingly,
the learned stiffness in $y$ direction became slightly higher ($474$ N/m)
compared to its initial value. During physical interaction the forces acting
along the y-axis are mostly responsible for supporting the weight of the
object.
With a higher stiffness value interaction times became lower and also helped 
avoiding situations where the robot did not release the object. The learned
slope 
parameter of the finger controller became more conservative ($3.63$ 1/N). 
This prevented any finger movement until the human force reached at least
$0.8\times$ the weight of the object. Finally, the learned minimal and maximal
tracking distance on expectation became $269$ and $541$mm respectively. 
\begin{figure}[t]
\centering
\captionsetup{width=.8\linewidth}
\includegraphics[width = 0.9\textwidth]{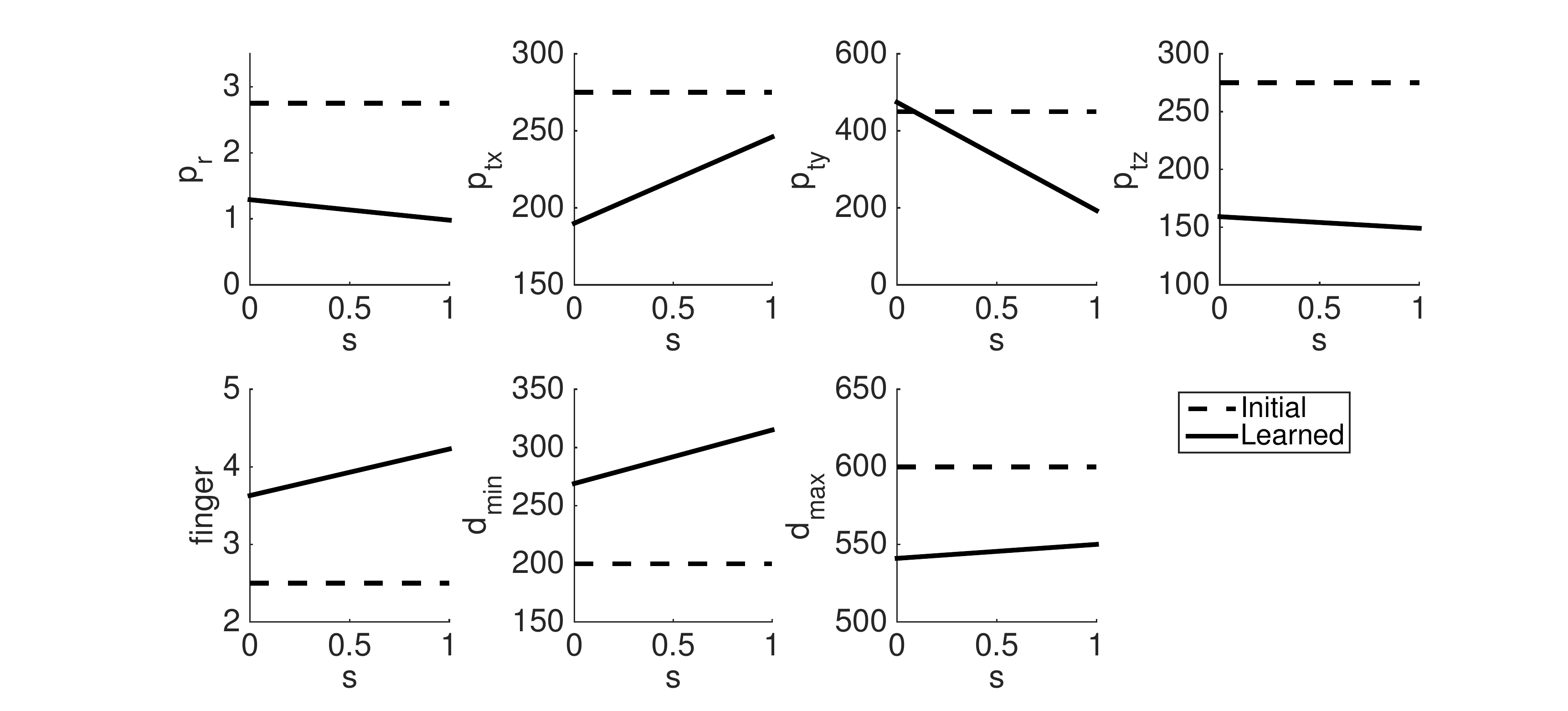}
\caption{The initial and the learned policy parameters against the context value.
Top row, from left to right: the rotational stiffness, translational stiffness in the
x-y-z direction. Bottom row, from left to right: finger control slope, minimal and maximal visual hand tracking
distance.}
\label{fig:parameters}
\end{figure}

The policy generalizes the controller parametrization with mean $\bs{a} +
\bs{A}\s$. We discussed above how $\bs{a}$ changed on expectation after the
learning. We now turn our attention to $\bs{A}$ and show how generalization to
more dynamic task happens. We typically executed experiments with hand speed between $0.1$ and
$1$m/s. We observed that 
on expectation the rotational stiffness values were lowered for more dynamical
tasks ($\s =1$m/s) with $-0.31$ Nm/rad. This helped the robot to orient with the
human hand quicker. Interestingly, we observed that the stiffness in x direction
is slightly increased with $56$ N/m. However, the stiffness in $y$ direction is
dramatically decreased with $-281$ N/m. This reduces forces acting on the human
significantly during faster physical interaction. The stiffness in 
$z$ direction is decreased with $-10$ N/m, which is just a minor
change. Interestingly, the slope parameter of the robot finger controller
increases with $0.6$ 1/N, which leads to an even more conservative finger
control. Finally, we observed that on expectation the minimal hand tracking
distance is increased by $46$mm and the maximal distance remains almost the same
with an additional $9$mm. A visual
representation of the learned parameters against the
context value is shown in Figure \ref{fig:parameters}.  
In the following, we will analyze some static and dynamic
handover experiments to give more insight why humans prefer the learned policy
as opposed to the initial hand-tuned controller. 

\begin{figure}
\centering
\subfigure[]{
\captionsetup{width=.8\linewidth}
 \label{fig:static}
 \includegraphics[width = .4\textwidth]{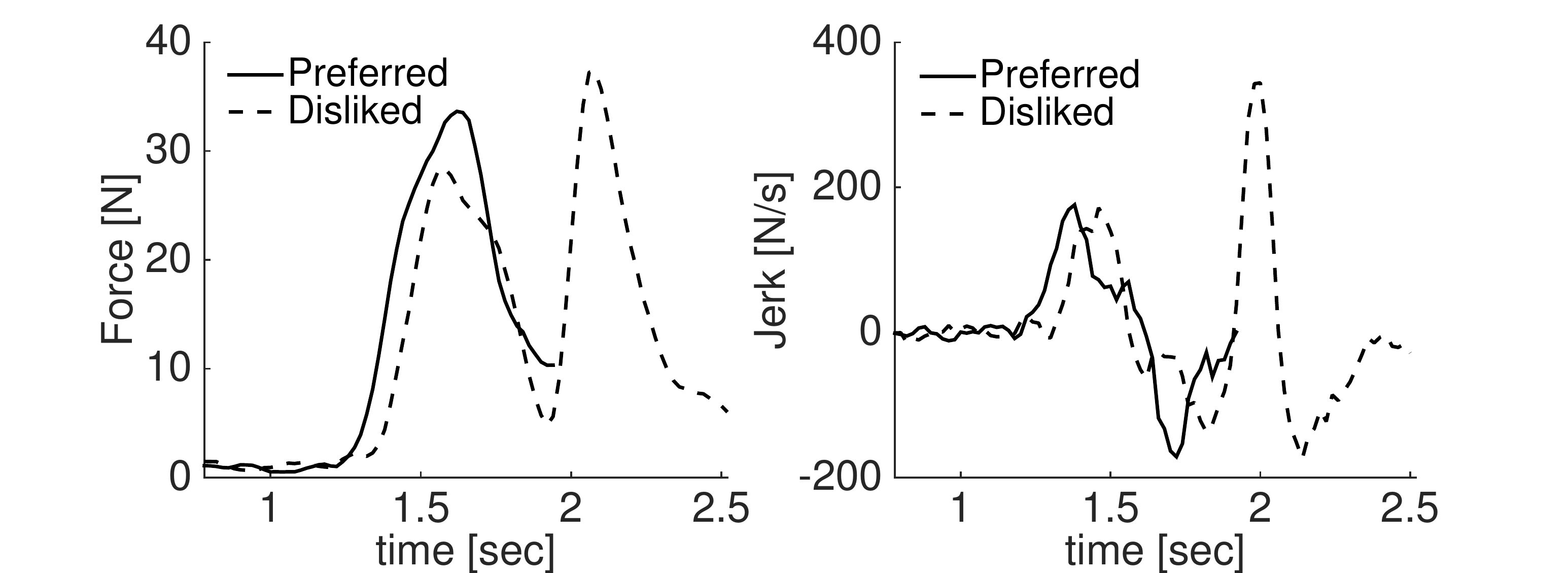}
 }\quad
 \subfigure[]{
 \label{fig:dynamic}
 \includegraphics[width = .4\textwidth]{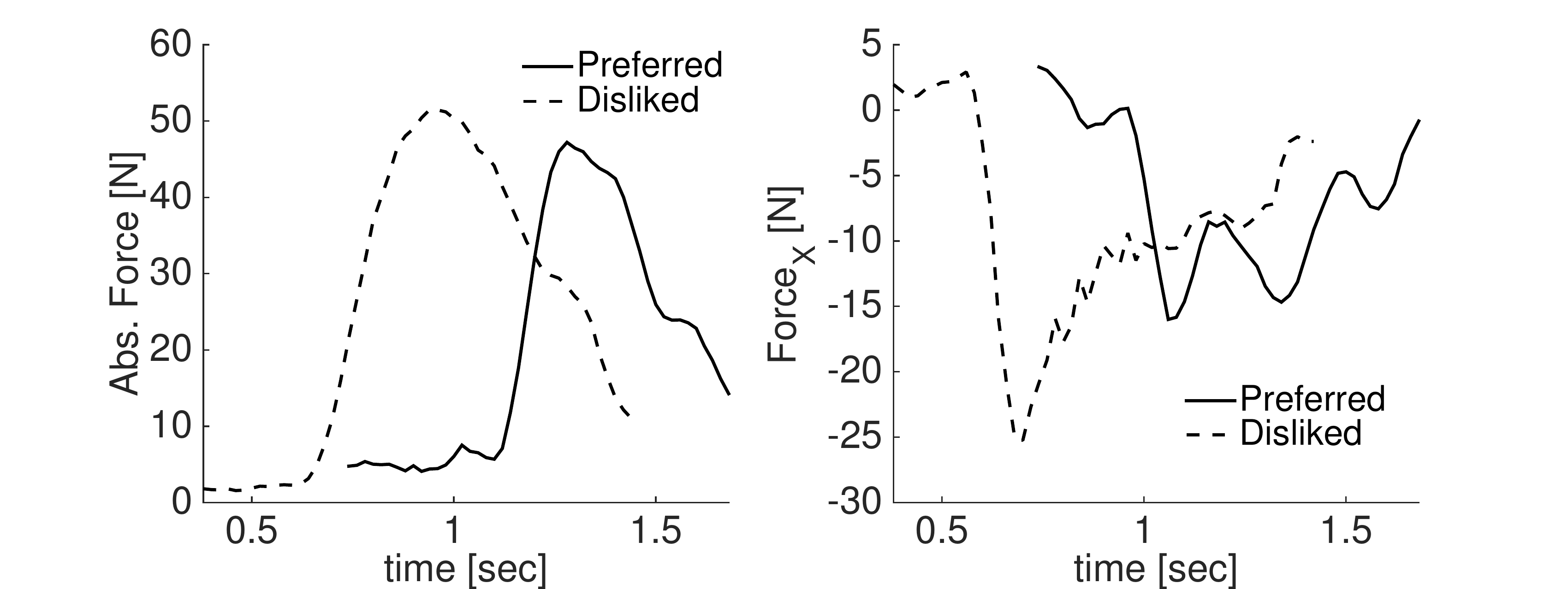}
 }
\captionsetup{width=.8\linewidth}
 \caption{\textbf{(a)} Two example of experimental results of the forces acting between the
human and the robot during physical interaction. The forces are plotted starting
right before the physical interaction until the handover is finished.
\textbf{(b)}  Two example of experimental results in dynamic handover situations. The forces are plotted starting
right before the physical interaction until the handover is finished.}
 \end{figure}

\textbf{Human preferences for static handovers.} For static handover tasks we
observed that a 
robust and quick finger control was always preferred and highly rated. In Figure
\ref{fig:static} we can see the forces and jerks of two typical static handover
solutions. The weight of the bottle was around $20$N. We can see that the
preferred solution always maintained a low jerk and forces remained limited.
Moreover, a successful handover happens relatively fast. In our experiments we
observed that a high quality solution happens within $0.6$ seconds and no faster
than $0.4$ seconds. Similar results have been reported in human-human object
transfers experiments \citep{ChaPar13}. Typically disliked
parameterizations have low translational stiffness and a stiff finger control,
resulting in the robot not releasing the object quick enough, which is
considered a failure. These experiments typically lasted for $1$ to $2$ seconds
until the bottle was released. 

\textbf{Human preferences for dynamic handovers.} In dynamic handover situations
contact forces and jerks were significantly higher compared to the static case
(Figure \ref{fig:dynamic}). A typical preferred dynamic handover controller has
lower rotational and translational stiffness,  and  a more firm finger controller. In our experiments
the human always came from one direction while taking the bottle from the robot.
In the robot hand frame this was the x-direction. As we can see, a preferred
controller achieves a significantly lower contact force and jerk in this
direction. We noticed that a physical contact time in a
dynamic handover  scenario is around $0.3-0.6$ sec. Based on the latent rewards,
we noticed that there is a strong preference towards faster handovers, as
opposed to the static case, where we did not observe such strong correlation in
handovers within $0.6$ seconds. Interestingly,  we noticed that humans preferred
stiffer finger controllers in dynamic handovers. We assume that
this helps a robust transfer of the object from giver to taker. In a
dynamic handover situation vision might not provide enough feedback about
the handover situation during physical contact, and thus, an excess of grip force would be necessary to
ensure the robust transfer and to compensate for inaccurate position control. 
 Video footage of some typical experiments before and after the learning is
 available at   www.youtube.com/watch?v=2OAnyfph3bQ.

 By analyzing these experiments we can see that the learned policy indeed
 provides a controller parametrization that decreases handover time, reduces
 forces and jerks acting on the human over a wide variety of dynamic situations.
 While the initial policy provides a reasonable performance in less dynamic
 experiments, learning and generalization significantly improves the performance
 of the policy. Based on our observations, for static handovers a fast and smooth finger
 control was necessary for success, while in dynamic handover situation higher
 compliance and a firm finger control were preferred. 

\section{Discussion}
This paper presents an algorithm for learning dynamic robot-to-human object
handover from human feedback. The algorithm learns a latent reward function
from both absolute and preference feedback, and integrates reward learning
with contextual policy search.  Experiments show that the robot adapts to the
dynamics of human motion and learns to hand over a water bottle successfully,
even in highly dynamic situations.

The current work has several limitations.  First, it is evaluated on a single
object and a small number of people.  We plan to generalize the learning
algorithm to adapt over human preferences and object characteristics.  While
contextual policy search works well for adapting over handover dynamics,
object characteristics exhibit much greater variability and may pose greater
challenge.  Second, our handover policy also does not consider human response
during the handover or its change over time.  We want to model key features of
human response and exploit it for effective and fluid handover. For both,
combining model-free learning and model-based planning seems a fruitful
direction for exploration.

\bigskip
\noindent\textbf{Acknowledgements.}
This research was supported in part an A*STAR Industrial Robotics Program
grant (R-252-506-001-305) and a SMART Phase-2 Pilot grant (R-252-000-571-592). 

\bibliographystyle{plain}

\end{document}